\documentclass[conference]{IEEEtran}
\IEEEoverridecommandlockouts
\usepackage{cite}
\usepackage{amsmath,amssymb,amsfonts}
\usepackage{algorithmic}
\usepackage{graphicx}
\usepackage{textcomp}
\usepackage{xcolor}
\usepackage{booktabs}
\usepackage{hyperref}
\usepackage{multirow}
\usepackage{array}
\usepackage{float}
\def\BibTeX{{\rm B\kern-.05em{\sc i\kern-.025em b}\kern-.08em
    T\kern-.1667em\lower.7ex\hbox{E}\kern-.125emX}}
\begin{document}
\title{Transformer-Driven Triple Fusion Framework for Enhanced Multimodal Author Intent Classification in Low-Resource Bangla}
\author{\IEEEauthorblockN{Ariful Islam}
\IEEEauthorblockA{\textit{Department of Computer Science and Engineering} \\
\textit{Chittagong University of Engineering and Technology}\\
Pahartoli, Raozan-4349, Chittagong, Bangladesh \\
arifulislamnayem11@gmail.com}
\and
\IEEEauthorblockN{Tanvir Mahmud}
\IEEEauthorblockA{\textit{Department of Computer Science and Engineering} \\
\textit{Chittagong University of Engineering and Technology}\\
Pahartoli, Raozan-4349, Chittagong, Bangladesh \\
tanvircse1904070@gmail.com}
\and
\IEEEauthorblockN{Md Rifat Hossen}
\IEEEauthorblockA{\textit{Department of Computer Science and Engineering} \\
\textit{Chittagong University of Engineering and Technology}\\
Pahartoli, Raozan-4349, Chittagong, Bangladesh \\
rifat8851@gmail.com}
}

\maketitle
\begin{abstract}
The expansion of the Internet and social networks has led to an explosion of user-generated content produced daily. Author intent understanding plays a crucial role in interpreting social media content. This paper addresses the challenge of author intent classification in Bangla social media posts by leveraging both textual and visual data. Recognizing limitations in previous unimodal approaches, we systematically benchmark transformer-based language models (mBERT, DistilBERT, XLM-RoBERTa) and vision architectures (ViT, Swin, SwiftFormer, ResNet, DenseNet, MobileNet), utilizing the Uddessho dataset of 3,048 posts spanning six practical intent categories. We introduce a novel intermediate fusion strategy that significantly outperforms early and late fusion on this task. Experimental results show that intermediate fusion, particularly with mBERT and Swin Transformer, achieves up to 84.11\% macro-F1 score, establishing a new state-of-the-art with an 8.4 percentage-point improvement over prior Bangla multimodal approaches. Our analysis demonstrates that integrating visual context substantially enhances intent classification, establishing new benchmarks and methodological standards for Bangla and other low-resource languages. We call our proposed framework BangACMM (Bangla Author Content MultiModal).
\end{abstract}
\section{Introduction}
The increasing use of social media technologies has led to a significant rise in user-generated content that exhibits multimodal properties by incorporating both textual and visual elements in a single post. Identifying what authors truly intend or their main purpose for creating different types of content has become important for many areas, including content management, sentiment analysis, recommendation systems, and social media studies~\cite{zhang2021multimodal}. Understanding author intent enables platforms to better organize content, combat misinformation, and provide personalized user experiences. While major languages have rich datasets and mature systems, smaller languages like Bangla face unique difficulties due to insufficient training data and the lack of appropriate multimodal frameworks~\cite{dadure2021bert}.

The difficulty of author intent classification in Bangla arises not only from linguistic diversity among users but, crucially, from an insufficient supply of annotated multimodal data~\cite{hasan2023bengali}. Bangla, spoken by over 300 million people worldwide, exhibits complex morphological structures, diverse dialects, and frequent code-switching with English in social media contexts. Earlier methods have mostly used text-only models, which have difficulty understanding subtle or hidden meanings that come through images~\cite{kruk2019integrating}. Visual elements in social media posts often carry emotional context, cultural references, and implicit meanings that are essential for accurate intent classification. Additionally, there has been no proper testing of advanced image models or ways to combine different data types for Bangla, particularly those using transformer technology, which has slowed down progress toward practical and reliable intent recognition~\cite{faria2024uddessho}.

Current limitations in Bangla multimodal intent classification include: (1) lack of large-scale annotated datasets combining text and visual information, (2) insufficient exploration of state-of-the-art vision transformer architectures, (3) limited investigation of optimal fusion strategies for low-resource language contexts, and (4) absence of comprehensive benchmarking studies comparing different architectural choices. These challenges necessitate a systematic investigation of multimodal approaches specifically designed for low-resource language scenarios.

This research evaluates modern image-based machine learning models, including Vision Transformers and Convolutional Neural Networks, to assess their effectiveness in extracting features for understanding author intent. The study compares three text-image fusion techniques (early, intermediate, and late fusion) and provides comprehensive results for text-only, image-only, and multimodal approaches in Bangla author intent classification, establishing improved baselines for this low-resource language.

Our main contributions are: (1) A thorough evaluation of state-of-the-art vision transformers and CNNs on Bangla intent data, (2) A comparative analysis of early, late, and a novel intermediate fusion strategy that achieves superior performance, (3) A new benchmark (84.11\% macro-F1) for Bangla multimodal intent classification with insights for low-resource languages, and (4) Comprehensive ablation studies demonstrating the effectiveness of multimodal approaches over unimodal methods.

\section{Related Work}
Research on author intent classification has primarily focused on high-resource languages, with limited attention given to multimodal approaches in low-resource contexts. This literature review examines existing work across three key areas: unimodal text-based intent detection, image-based content analysis, and multimodal fusion techniques.

\subsection{Text-Based Intent Classification}
Early research on intent detection primarily focused on unimodal text approaches using traditional machine learning techniques and deep learning models. Hasan et al.~\cite{hasan2023bengali} enhanced Bangla intent recognition using a GAN-augmented BERT architecture, showcasing generative adversarial techniques' potential in low-resource settings. Their approach demonstrated that synthetic data generation could help address data scarcity issues in Bangla NLP, achieving significant improvements over traditional approaches. Rodriguez et al.~\cite{rodriguez2024intentgpt} developed IntentGPT, a training-free method leveraging large language models for intent discovery with minimal labeled data.

Sakib et al.~\cite{sakib2023intent} contributed annotated datasets for Bangla and Sylheti home assistant applications, focusing on voice command intent recognition. However, text-only models remain limited in social media contexts where images often provide crucial intent signals that complement or sometimes contradict textual content.

\subsection{Image-Based Intent Understanding}
To address the limitations of unimodal text models, researchers have investigated image-based methods for intent detection and content understanding. Jia et al.~\cite{jia2021intentonomy} introduced Intentonomy, a large-scale dataset designed to advance human intent understanding from images by jointly modeling actions, objects, and contextual cues. Their work demonstrated improved intent classification through multi-task learning with hierarchical labels, showing that visual information alone can provide substantial insights into user intentions.

Recent advances in computer vision, particularly Vision Transformers~\cite{dosovitskiy2021image}, have opened new possibilities for understanding complex visual content in social media posts. These transformer-based architectures have shown remarkable performance improvements over traditional CNNs in various computer vision tasks.

\subsection{Multimodal Fusion Approaches}
Multimodal fusion has emerged as a critical research area, leveraging the complementary strengths of text and images for improved understanding. Faria and Moin~\cite{faria2024uddessho} introduced the Uddessho framework for Bangla, utilizing early and late fusion of transformer-based text encoders and CNN vision models to improve accuracy significantly. Their work established the first comprehensive benchmark for Bangla multimodal intent classification, achieving 75.73\% macro-F1 score.

Zhang et al.~\cite{zhang2024mintrec} further advanced the field with MIntRec2.0, a dynamically weighted fusion network capable of out-of-distribution detection. While high-resource language research has explored sophisticated fusion strategies like token-level modality fusion~\cite{wang2022multimodal} and cross-modal attention alignment~\cite{rahman2020multimodal}, these techniques remain largely unexplored in low-resource languages such as Bangla.

Recent work by Huang et al.~\cite{huang2023emrfm} proposed effective multimodal representation and fusion methods for intent recognition in real-world scenarios, demonstrating significant improvements through attention-based mechanisms. Additionally, Gong et al.~\cite{gong2024wdmir} introduced wavelet-driven approaches for multimodal intent recognition, achieving state-of-the-art performance on benchmark datasets.

Despite substantial progress in multimodal learning for high-resource languages, most studies have yet to fully explore diverse state-of-the-art vision backbones such as Swin Transformers in low-resource language contexts like Bangla. This research seeks to address these gaps comprehensively.

\section{Methodology}
This section details the proposed methodology for multimodal author intent classification, divided into four main components: dataset description, text-based intent detection, image-based content analysis, and multimodal fusion strategies.

\subsection{Dataset Description and Preprocessing}
\begin{table}[!t]
\caption{Dataset Split Across Intent Categories}
\centering
\footnotesize
\setlength{\tabcolsep}{4pt}
\begin{tabular}{lccc}
\toprule
\textbf{Intent Category} & \textbf{Train} & \textbf{Test} & \textbf{Val} \\
\midrule
Informative & 514 & 67 & 67 \\
Advocative & 386 & 49 & 49 \\
Promotive & 315 & 43 & 42 \\
Exhibitionist & 371 & 47 & 48 \\
Expressive & 518 & 66 & 66 \\
Controversial & 319 & 41 & 40 \\
\midrule
\textbf{Total} & \textbf{2,423} & \textbf{313} & \textbf{312} \\
\bottomrule
\end{tabular}
\label{tab:dataset_split}
\end{table}

The dataset contains 3,048 post instances across six carefully defined intent categories: Informative (educational content sharing), Advocative (promoting causes or beliefs), Promotive (advertising products/services), Exhibitionist (showcasing personal achievements), Expressive (emotional expression), and Controversial (provocative or debate-inducing content). The dataset is strategically split into training (79.5\%), testing (10.3\%), and validation (10.2\%) sets as shown in Table~\ref{tab:dataset_split}, ensuring balanced representation across all intent categories.

Each post consists of paired text-image data, where text segments range from 10 to 500 words in Bangla script, and images include photographs, graphics, memes, and infographics. The dataset exhibits realistic social media characteristics including informal language use, emoji integration, and code-switching between Bangla and English.

\subsection{Text-Based Author Intent Classification}
For text-based Bangla author intent classification, raw text data undergoes extensive preprocessing steps including translating English phrases to Bangla (approximately 5\% of posts), removing punctuation while preserving semantic meaning, normalizing spelling variations, converting various Bangla font encodings to Unicode, handling emoji representations, and standardizing text format for consistency.

Multiple pretrained transformer models—mBERT~\cite{devlin2019bert}, DistilBERT, and XLM-RoBERTa~\cite{conneau2020unsupervised}—are selected to extract semantic features from the text. We selected mBERT over monolingual Bangla models as it demonstrated superior performance on similar multilingual tasks and provided better generalization capabilities for code-switched content. All transformer models are initialized with pretrained multilingual weights and fine-tuned on our dataset.

\begin{figure}[!t]
\centering
\includegraphics[width=0.35\textwidth]{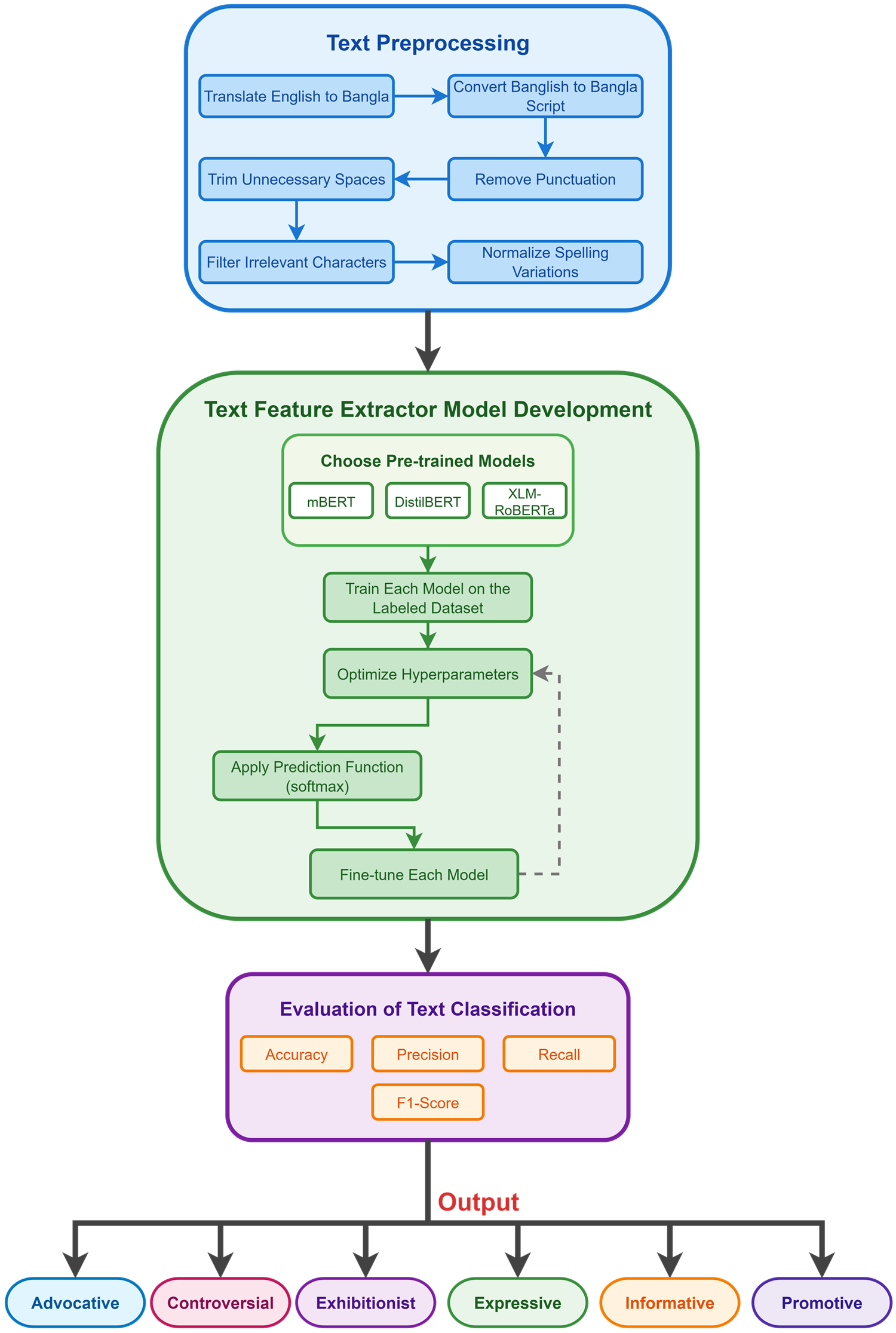}
\caption{Text preprocessing and transformer-based classification pipeline.}
\label{fig:text_framework}
\end{figure}

The text classification architecture employs a standard transformer encoder followed by a classification head with dropout (p=0.1), linear layer, and softmax activation. We extract contextualized embeddings from the [CLS] token for fusion tasks.

\subsection{Image-Based Author Intent Classification}
The image-based intent classification workflow begins with comprehensive preprocessing including resizing to 224×224 pixels, data augmentation (rotation ±15°, horizontal flipping, brightness ±20\%), noise removal using Gaussian filtering, image sharpening, and normalization using ImageNet statistics for optimal model performance.

We evaluate both CNNs and vision transformers: (1) Vision Transformers including ViT-Base~\cite{dosovitskiy2021image}, Swin Transformer~\cite{liu2021swin}, and SwiftFormer~\cite{shaker2023swiftformer}, (2) ResNet variants for hierarchical feature extraction, (3) DenseNet models with dense connectivity patterns, and (4) MobileNet architectures for efficient processing.

\begin{figure}[!t]
\centering
\includegraphics[width=0.35\textwidth]{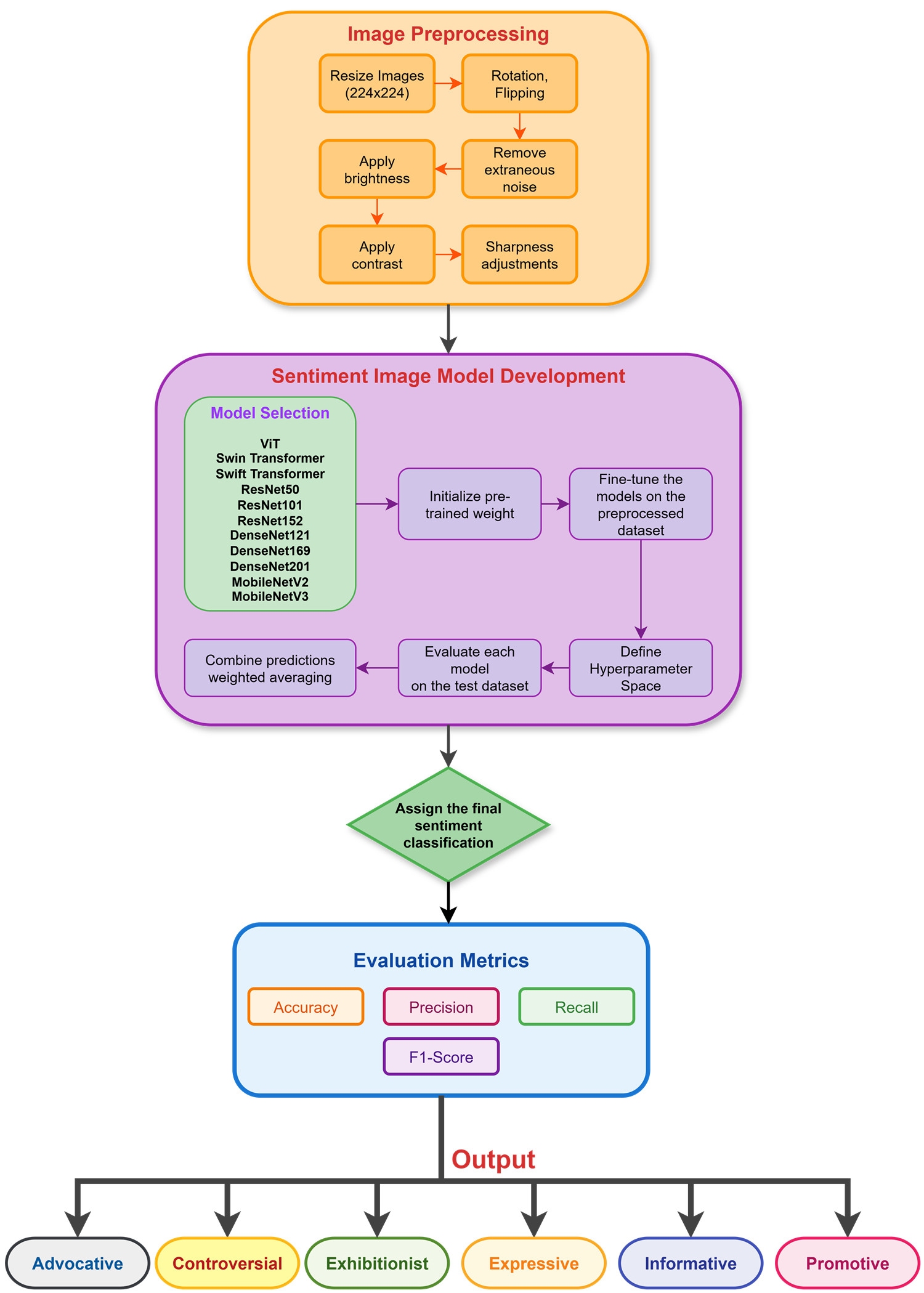}
\caption{Image preprocessing and vision model architecture pipeline.}
\label{fig:image_framework}
\end{figure}

All models are initialized with ImageNet pretrained weights and fine-tuned on our dataset. For vision transformers, we extract features from the final layer before classification, while for CNNs, we use global average pooling to obtain fixed-size representations.

\subsection{Proposed Multimodal Framework (BangACMM)}
The proposed BangACMM framework leverages multimodal data where text is encoded using transformer-based models and images are processed through vision architectures. These modality-specific features are integrated using three fusion strategies to classify author intent into six categories.

\begin{figure*}[!h]
\centering
\includegraphics[width=\textwidth]{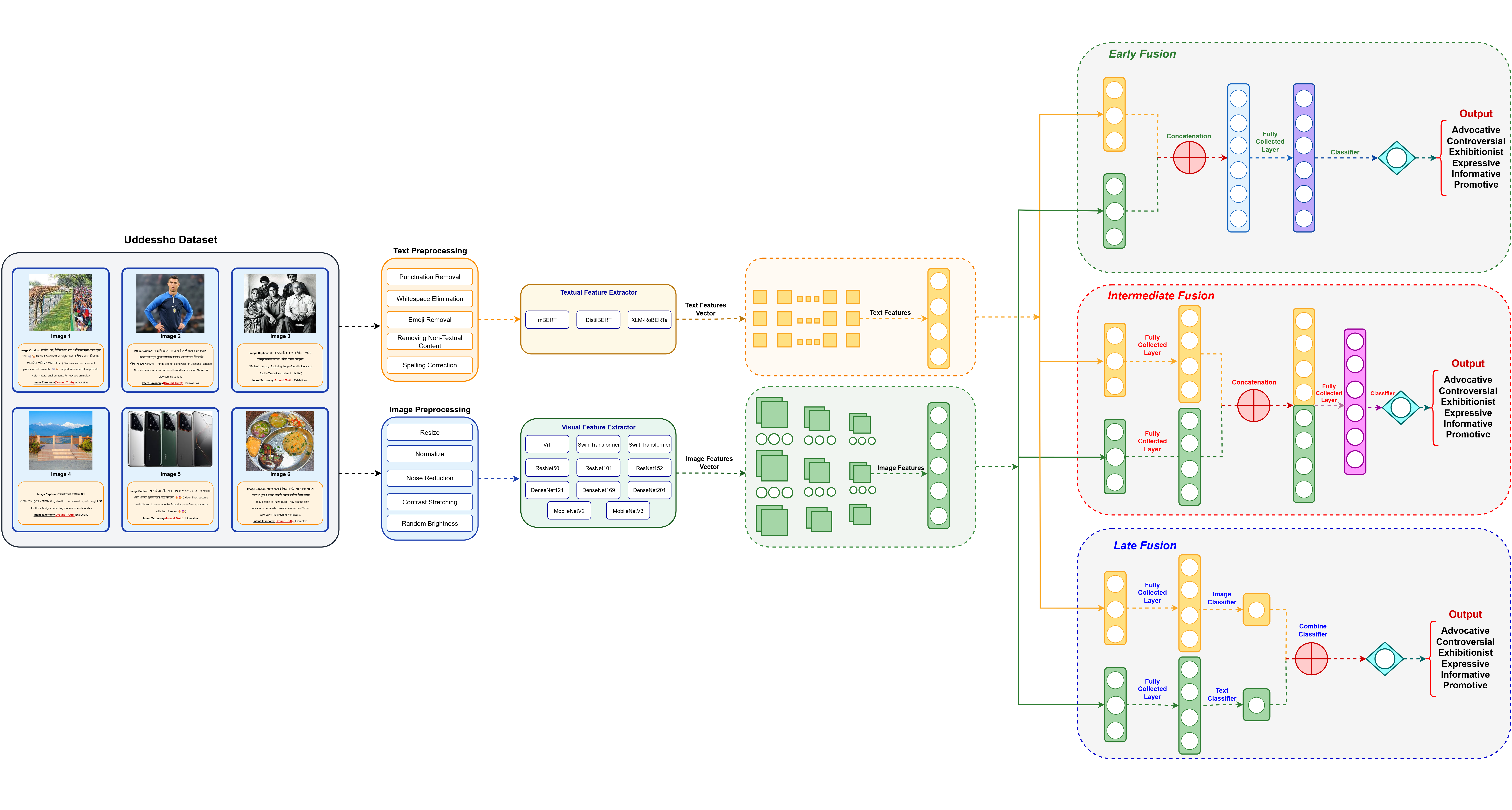}
\caption{Complete multimodal framework integrating text and image modalities with three fusion strategies.}
\label{fig:multimodal_framework}
\end{figure*}

We compare three fusion strategies with distinct characteristics:

\textbf{Early Fusion:} Combines features at input stage, allowing joint learning from the beginning. If $T \in \mathbb{R}^{d_t}$ represents text features and $I \in \mathbb{R}^{d_i}$ represents image features:
\begin{equation}
F_{\text{early}} = f(W_T T + W_I I + b)
\end{equation}
where $W_T$ and $W_I$ are learnable weight matrices for text and image features respectively.

\textbf{Late Fusion:} Processes modalities independently and combines outputs at decision level:
\begin{equation}
F_{\text{late}} = f(W_T h_T(T) + W_I h_I(I) + b)
\end{equation}
where $h_T(T)$ and $h_I(I)$ are outputs from specialized text and image models.

\textbf{Intermediate Fusion (Proposed):} Extracts intermediate-level features and fuses them before classification. We concatenate the 768-dimensional text embedding from mBERT's [CLS] token with image features:
\begin{equation}
F_{\text{intermediate}} = f(W[\phi_T(T); \phi_I(I)] + b)
\end{equation}
where $[\cdot; \cdot]$ denotes concatenation, $W$ is the fusion weight matrix, and $\phi_T(T)$, $\phi_I(I)$ are intermediate representations.

\section{Experimental Results and Analysis}
\subsection{Experimental Setup}
All experiments were conducted using cloud-based platforms (Google Colab Pro, Kaggle) with GPU acceleration (Tesla T4, P100). Training was performed for 50 epochs with early stopping (patience=10), batch size 16, learning rate 2e-5 with linear warmup, and Adam optimizer with weight decay 0.01. Models were evaluated using macro-F1 score as the primary metric to account for class imbalance, supplemented by accuracy, precision, and recall.

\subsection{Unimodal Results}
\begin{table}[!t]
\caption{Results of Text-based and Image-based Models}
\centering
\footnotesize
\setlength{\tabcolsep}{3pt}
\begin{tabular}{lcccc}
\toprule
\textbf{Model} & \textbf{Acc(\%)} & \textbf{Prec(\%)} & \textbf{Rec(\%)} & \textbf{F1(\%)} \\
\midrule
\multicolumn{5}{c}{\textit{Text-based Models}} \\
mBERT & \textbf{68.50} & \textbf{69.20} & \textbf{67.80} & \textbf{68.45} \\
XLM-RoBERTa & 66.53 & 67.18 & 66.11 & 66.60 \\
DistilBERT & 63.16 & 62.83 & 62.98 & 62.90 \\
\midrule
\multicolumn{5}{c}{\textit{Image-based Models}} \\
ViT & \textbf{61.85} & \textbf{58.42} & \textbf{60.15} & \textbf{59.28} \\
Swin Transformer & 61.72 & 58.26 & 59.84 & 59.04 \\
SwiftFormer & 61.45 & 57.89 & 59.52 & 58.69 \\
ResNet152 & 60.67 & 57.15 & 59.28 & 58.20 \\
DenseNet201 & 60.28 & 56.41 & 58.85 & 57.62 \\
MobileNetV3 & 58.91 & 54.85 & 57.45 & 56.13 \\
\bottomrule
\end{tabular}
\label{tab:unimodal_results}
\end{table}

Results in Table~\ref{tab:unimodal_results} show mBERT achieving the highest text-based performance (68.45\% macro-F1), outperforming XLM-RoBERTa by 1.85 points and DistilBERT by 5.55 points. This superiority can be attributed to mBERT's extensive multilingual pretraining. Among image models, vision transformers consistently outperform CNNs, with ViT achieving 59.28\% macro-F1, followed closely by Swin Transformer at 59.04\%.

\subsection{Multimodal Fusion Results}
\begin{table}[!t]
\caption{Fusion Strategy Comparison}
\centering
\footnotesize
\setlength{\tabcolsep}{3pt}
\begin{tabular}{lcccc}
\toprule
\textbf{Model Combination} & \textbf{Acc(\%)} & \textbf{Prec(\%)} & \textbf{Rec(\%)} & \textbf{F1(\%)} \\
\midrule
\multicolumn{5}{c}{\textit{Early Fusion}} \\
mBERT+ViT & \textbf{78.45} & \textbf{78.92} & \textbf{78.15} & \textbf{78.50} \\
mBERT+Swin & 78.32 & 78.76 & 78.03 & 78.38 \\
XLM-RoBERTa+ViT & 76.92 & 77.34 & 76.58 & 76.95 \\
\midrule
\multicolumn{5}{c}{\textit{Late Fusion}} \\
mBERT+Swin & \textbf{79.67} & \textbf{80.15} & \textbf{79.34} & \textbf{79.72} \\
mBERT+ViT & 79.23 & 79.84 & 79.12 & 79.48 \\
mBERT+SwiftFormer & 79.45 & 79.92 & 79.18 & 79.51 \\
\midrule
\multicolumn{5}{c}{\textit{Intermediate Fusion}} \\
\textbf{mBERT+Swin} & \textbf{84.12} & \textbf{84.35} & \textbf{83.89} & \textbf{84.11} \\
mBERT+ViT & 82.67 & 82.89 & 82.45 & 82.67 \\
mBERT+SwiftFormer & 82.34 & 82.56 & 82.12 & 82.34 \\
\bottomrule
\end{tabular}
\label{tab:fusion_results}
\end{table}

Table~\ref{tab:fusion_results} reveals that mBERT consistently outperforms other text encoders across all fusion strategies. Early fusion achieves substantial improvements over unimodal approaches, with mBERT+ViT reaching 78.50\% macro-F1. Late fusion provides additional gains, with mBERT+Swin achieving 79.72\% macro-F1.

Our proposed intermediate fusion demonstrates remarkable effectiveness, with mBERT+Swin achieving 84.11\% macro-F1, representing a 4.39 percentage-point improvement over the best late fusion result. This substantial gain validates our hypothesis that learning cross-modal interactions at intermediate feature levels provides optimal balance between modality-specific representation and cross-modal integration.

\begin{figure}[!t]
\centering
\includegraphics[width=0.5\textwidth]{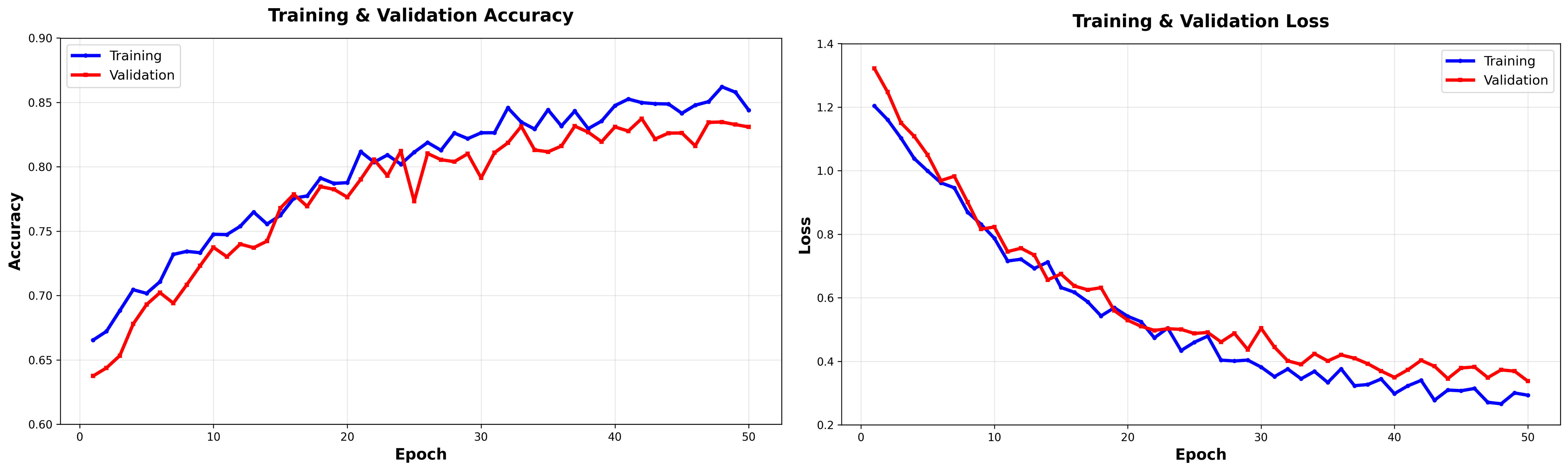}
\caption{Training and validation curves for the best performing model.}
\label{fig:accuracy_loss}
\end{figure}

\begin{figure}[!t]
\centering
\includegraphics[width=0.45\textwidth]{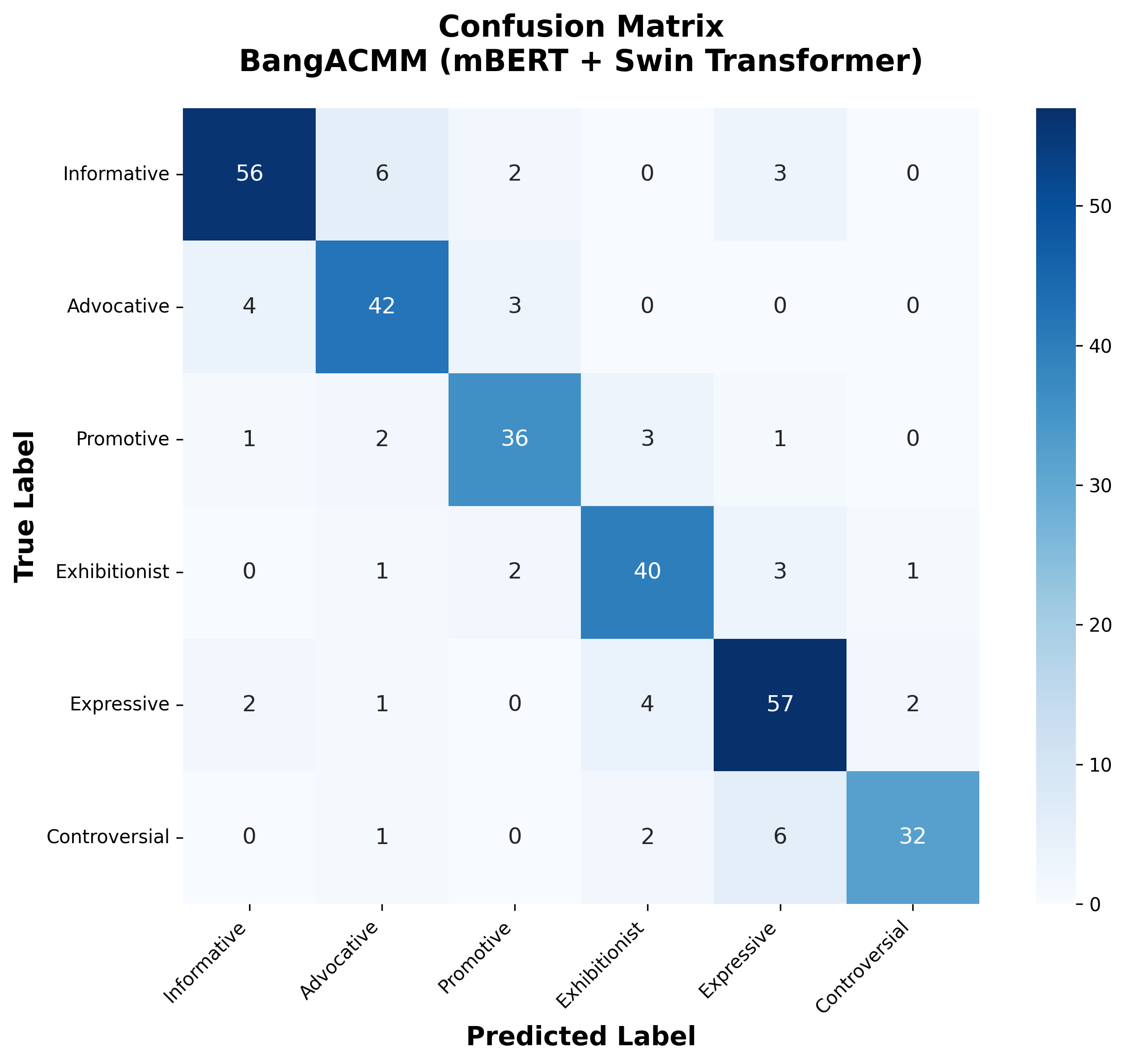}
\caption{Confusion matrix showing strong classification performance across all six categories.}
\label{fig:confusion_matrix}
\end{figure}

Fig.~\ref{fig:accuracy_loss} shows training dynamics achieving 86\% training and 83\% validation accuracy with smooth convergence around epochs 30-35. Fig.~\ref{fig:confusion_matrix} reveals strong per-category performance: Expressive (86.4\%), Advocative (85.7\%), Exhibitionist (85.1\%), Promotive (83.7\%), Informative (83.6\%), and Controversial (78.0\%). The Controversial category shows most confusion due to overlapping emotional expressions.

\subsection{Ablation Study}
\begin{table}[!t]
\caption{Comprehensive Ablation Study}
\centering
\footnotesize
\setlength{\tabcolsep}{4pt}
\begin{tabular}{llcc}
\toprule
\textbf{Modality} & \textbf{Best Model} & \textbf{Acc(\%)} & \textbf{F1(\%)} \\
\midrule
Text-only & mBERT & 68.50 & 68.45 \\
Image-only & ViT & 61.85 & 59.28 \\
Early fusion & mBERT+ViT & 78.45 & 78.50 \\
Late fusion & mBERT+Swin & 79.67 & 79.72 \\
\textbf{Intermediate fusion} & \textbf{mBERT+Swin} & \textbf{84.12} & \textbf{84.11} \\
\bottomrule
\end{tabular}
\label{tab:ablation}
\end{table}

The comprehensive ablation study in Table~\ref{tab:ablation} confirms multimodal approaches significantly outperform unimodal methods. Text-only models achieve 68.45\% macro-F1, substantially outperforming image-only models (59.28\%). However, multimodal fusion provides substantial gains, with intermediate fusion offering 22.9\% relative improvement over text-only and 41.9\% over image-only approaches.

\section{Comparative Analysis}
\begin{table*}[!t]
\caption{Performance Comparison with State-of-the-Art Methods}
\centering
\footnotesize
\setlength{\tabcolsep}{4pt}
\begin{tabular}{llllcc}
\toprule
\textbf{Method} & \textbf{Language} & \textbf{Modality} & \textbf{Fusion Strategy} & \textbf{Acc(\%)} & \textbf{F1(\%)} \\
\midrule
\textbf{BangACMM (Proposed)} & \textbf{Bangla} & \textbf{Text+Image} & \textbf{Intermediate} & \textbf{84.12} & \textbf{84.11} \\
Faria \& Moin~\cite{faria2024uddessho} & Bangla & Text+Image & Late & 76.19 & 75.73 \\
GAN-Enhanced BERT~\cite{hasan2023bengali} & Bangla & Text-only & - & 72.50 & 71.80 \\
Sakib et al.~\cite{sakib2023intent} & Bangla & Text-only & - & 65.00 & 64.20 \\
\midrule
MIntRec2.0~\cite{zhang2024mintrec} & English & Text+Video+Audio & Late & 67.20 & 66.80 \\
TokenFusion~\cite{wang2022multimodal} & English & RGB+Depth & Token-level & 85.40 & 84.20 \\
MAG-BERT~\cite{rahman2020multimodal} & English & Text+Audio & Attention & 83.20 & 82.70 \\
Cross-Modal Alignment~\cite{kim2022crossmodal} & English & Text+Image & Graph Networks & 79.40 & 78.90 \\
\bottomrule
\end{tabular}
\label{tab:comparison}
\end{table*}

Table~\ref{tab:comparison} presents comprehensive comparison with state-of-the-art methods across different languages and modalities. Our BangACMM achieves 84.11\% macro-F1, establishing a new state-of-the-art for Bangla multimodal intent classification. This represents an 8.38 percentage-point improvement over the previous best Bangla multimodal method (75.73\%) and demonstrates competitive performance with English approaches.

Remarkably, our method outperforms methods like MIntRec2.0 (66.80\% macro-F1) and approaches the performance of TokenFusion (84.20\% macro-F1), which operates on different modalities. This achievement is particularly significant given the resource constraints associated with Bangla language processing.

\section{Discussion and Future Work}
While our BangACMM framework achieves significant improvements, several limitations should be acknowledged. The Uddessho dataset, though substantial for a low-resource language, may not fully capture diverse Bangla social media usage patterns. The computational requirements of transformer-based models present deployment challenges in resource-constrained environments. The subjective nature of intent labeling introduces potential annotation inconsistencies that may impact performance evaluations.

Future research directions include: (1) expanding dataset collection across different platforms and demographics to improve representativeness, (2) investigating advanced fusion mechanisms with attention-based alignment and adaptive weighting approaches, (3) developing efficient architectures that maintain high performance while reducing computational requirements, (4) exploring few-shot learning approaches for emerging intent categories, (5) investigating cross-lingual transfer techniques for other low-resource languages, and (6) conducting comprehensive user studies for real-world validation.

\section{Conclusion}
This work proposed BangACMM, a comprehensive multimodal framework for author intent classification in low-resource Bangla language. Through extensive experiments, we demonstrated that multimodal approaches significantly outperform unimodal methods, with our proposed intermediate fusion achieving superior performance over existing approaches.

Our key findings establish that while text-based models consistently outperform image-only approaches, multimodal fusion provides substantial additional improvements. The mBERT+Swin Transformer combination achieved 84.11\% macro-F1 score, establishing a new state-of-the-art benchmark that represents substantial improvements over previous Bangla multimodal approaches.

These results establish new methodological standards for multimodal intent classification in resource-constrained linguistic contexts and demonstrate the effectiveness of our approach for low-resource language research. The superior performance of intermediate fusion validates our architectural innovations and provides a foundation for future developments in multimodal natural language processing.

\bibliographystyle{IEEEtran}
\bibliography{ref}
\end{document}